\documentclass[letterpaper, 10 pt, conference]{ieeeconf}  % Comment this line out if you need a4paper

\IEEEoverridecommandlockouts                              % This command is only needed if 
\usepackage[utf8]{inputenc} % allow utf-8 input
\usepackage[T1]{fontenc}    % use 8-bit T1 fonts
\usepackage{hyperref}       % hyperlinks
\usepackage{url}            % simple URL typesetting
\usepackage{booktabs}       % professional-quality tables
\usepackage{amsfonts}       % blackboard math symbols
\usepackage{nicefrac}       % compact symbols for 1/2, etc.
\usepackage{microtype}      % microtypography

\usepackage{amsmath,amssymb,mathrsfs}
\usepackage[ruled, linesnumbered, vlined]{algorithm2e}
\usepackage{comment}
\usepackage{graphicx}
\usepackage{graphics}
\usepackage[tight]{subfigure}
\usepackage{wrapfig}
\usepackage{float}
\usepackage[table]{xcolor}
\usepackage{tabu}
\usepackage{multirow}
\usepackage{array}
\usepackage{enumerate}
\usepackage{sidecap}
\usepackage{flushend}
\usepackage{hyperref}
\usepackage{url}
\graphicspath{{./figs/}}
\usepackage{changepage}   % for the adjustwidth environment

\newtheorem{definition}{\textbf{Definition}}[section]

\usepackage{color}

\newcommand{\vect}[1]{\mathbf{#1}}

\definecolor{lightblue}{rgb}{0,0.2,1}
\definecolor{black}{rgb}{0,0,0}
\hypersetup{
    colorlinks=true,%
    citebordercolor=white,%
    filebordercolor=white,%
    linkbordercolor=white,%
    urlbordercolor=white,%
    citecolor=black,%
    linkcolor=black,%
    urlcolor=black%
}

\newcounter{tecounter}
\DeclareMathOperator*{\argmax}{arg\,max}

\title{A Solution to Time-Varying Markov Decision Processes}

\author{Lantao Liu$^{1}$, Gaurav S. Sukhatme$^{2}$%
\thanks{$^{1}$Lantao Liu is with the Intelligent Systems Engineering Department at  Indiana University,
        Bloomington, IN 47408, USA. E-mail: {\tt\small lantao@iu.edu} }%
\thanks{$^{2}$Gaurav S. Sukhatme is with the Department of Computer Science at the University of Southern California, Los Angeles, CA 90089, USA. E-mail: {\tt\small gaurav@usc.edu}}\vspace{-5pt}
}

\begin{document}

\maketitle

\begin{abstract}
We consider a decision-making problem where the environment varies both in space and time. Such problems arise naturally when considering e.g., the navigation of an underwater robot amidst ocean currents or the navigation of an aerial vehicle in wind.
%problem with an underlying model (constructed from data) that describes spatiotemporal environmental dynamics.  
%A key challenge for such a decision-making framework to autonomous agents lies in that, the data obtained --- and therefore the model constructed --- can vary not only spatially, but also temporally.  
To model such spatiotemporal variation, we extend the standard Markov Decision Process (MDP) to a new framework called the Time-Varying Markov Decision Process (TVMDP). 
The TVMDP has a time-varying state transition model and transforms the standard MDP that considers only {\em immediate} and {\em static} uncertainty descriptions of state transitions, to a framework that is able to adapt to future time-varying transition dynamics over some horizon. 
We show how to solve a TVMDP via a redesign of the MDP value propagation mechanisms by incorporating the introduced dynamics along the temporal dimension.
We validate our framework in a marine robotics navigation setting using spatiotemporal ocean data 
and show that it outperforms prior efforts.
\end{abstract}

%%%%%%%%%%%%%%%%%%%%%%%%%%%%%%%%%%%%%%%%%%%%%%%%%%%%%%%%%%%%%%%%%%%%%%%%%%%%%%%%

%\vspace{-10pt}
\section{Introduction}
%\vspace{-5pt}

Consider a scenario where an underwater vehicle navigates across 
an area of ocean over a period of a few weeks to reach a goal location. 
The ocean currents are typically strong enough %(see Fig.~\ref{fig:ocean-currents}) 
to disturb the vehicle's motion, causing significantly uncertain action outcomes. Decision theoretic planning methods~\cite{BoutilierDTP99} cope with action uncertainty by stochastically modeling the action outcomes. However, the dynamic nature of the ocean currents implies that to be effective, any underlying model that describes the uncertainty associated with the vehicle's actions must also vary with time.

A popular method for addressing action uncertainty is the Markov Decision Process (MDP)~\cite{Bertsekas2000}, a decision-making framework in which the uncertainty due to actions is modeled using a stochastic state transition function. % (a probability distribution over resultant states when executing a particular action in a present state). 
However, a limitation of this approach is that the state transition model is static, i.e., the uncertainty distribution is a ``snapshot at a certain moment"~\cite{Puterman94MDP}.

Fortunately, environmental dynamics such as those associated with ocean currents can be forecast (albeit imprecisely) over a future time horizon. We exploit the idea that the forecast dynamics are time-varying functions that can be used to stochastically predict the state transition model. 
In this paper, we describe how to incorporate time-varying stochasticity into an MDP-style decision-making model and show how the resulting framework leads to improved planning accuracy. 

\textbf{Contributions: } 
 We propose a new method called the {\em time-varying Markov Decision Process (TVMDP)}. Different from the standard MDP which is designed for solving decision-making problems with time-invariant transition models,
the TVMDP has the capability of handling a time-varying transition model and can model decision problems with stochastic state transitions that vary both spatially and temporally. 
The proposed TVMDP method requires neither discretization of time nor augmentation of state space. But to solve it, we show that the basic Bellman backup mechanism used for computing the MDP is not enough. Instead, we present a new mechanism based on two iterative value propagation processes, which proceed in both spatial and temporal dimensions, where the spatial process involves Bellman backup but the temporal dimension is expanded using Kolmogorov equations.
This also implies that, the standard MDP is a special case of TVMDP.

\section{Related Works}

Time varying Markov transition dynamics have been studied previously in the context of pattern analysis of economic growth~\cite{time-varying-Morier11}, 
(aimed at understanding the dynamics of growth based on a collection of different states corresponding to different countries), analyzing fiscal policies~\cite{time-varing-agnello13} (civilian spending/taxing), and the environment~\cite{Hosseini12} (extreme temperature events). 
However, the models in these studies are Hidden Markov Models (HMMs), and unlike an MDP, they do not have the notion of an action to control transitions between states. 

The approach proposed in this paper bears comparison to a framework called the {\em time-dependent Markov Decision Process} (TDMDP)~\cite{Boyan00exactsolutions} that includes time in the state space of the MDP. In the TDMDP 
%focuses on analyzing the temporal effect of {\em stochastic travel duration} among two traversable states, i.e., 
the probability distribution associated with the transition model is used to describe the (uncertain) travel duration from one state to the next. This is similar to the conventional treatment of time-dependence in planning and routing in operations research~\cite{donati2008time}. In contrast, our approach (the TVMDP) follows the standard MDP convention and utilizes the transition model to describe the {\em uncertain transitions among states (stochastic state jumping),  while assuming that the transition model itself is time-varying}.
 The TDMDP uses a likelihood function to control the state transitions. In order to take advantage of the standard Bellman backup mechanism to compute the solution, a set of restrictions are imposed on almost every term of the TDMDP formulation, making the framework less applicable in many realistic scenarios. 

Recently, the TDMDP has been analyzed~\cite{RachelsonFG09}, and it has been shown that even under strong assumptions and restrictions, the computation can still be cumbersome. The same analysis recommends means for approximating a solution using strategies such as prioritized sweeping~\cite{moore1993prioritized}. The idea shares certain similarity to the {\em real-time dynamic programming}~\cite{bonet2003labeled} where those states with larger values or value changes have a higher priority to propagate. We use this approximation scheme on TDMDP to compare with our approach in the experiment section.
Differences between our approach and the TDMDP are compared more formally in the remainder of this work. 

Proximal works also include the vast literature on reinforcement learning (RL)~\cite{Sutton1998,DayanRL08,Bagnell_2013_7451} wherein an agent tries to maximize accumulated rewards by interacting with the environment during its lifetime, and the system is typically formulated as an MDP. A technique related to future prediction is temporal difference (TD) learning~\cite{Sutton88learningto,Tesauro1995TDL}. RL extends the TD technique by allowing the learned state-values to guide actions, and correct previous predictions based on availability of new observations~\cite{Sutton1998,Boyan99least-squarestemporal}. Unfortunately, existing TD techniques are based on time-invariant models. 

Additionally, in order to extend the discrete-time actions, a temporal abstraction based concept (the Semi-Markov Decision Processes (SMDP))~\cite{Sutton99betweenmdps} has been designed so that actions can be performed in a more flexible manner. SMDPs consider stochastic action durations, but still assume the transition model itself is time-invariant.

%%%%%%%%%%%%%%%%%%%%%%%%%%%%%%%%%%%%%%%%%%%%%%%%%%%%%%%%%%%%%%%%%%%%%%%%%%%%%%%%%%%%%

\section{Preliminaries}

\subsection{Markov Decision Process (MDP)}
\label{sec:prelim}

\begin{definition}
{\em
An MDP $\mathcal{M}$ is defined by a 4-tuple $\mathcal{M} = < S, A, T, R >$, where $S =\{s\}$ and $A=\{a\}$ represent the countable state space and action space, respectively. 
The stochastic {\em transition dynamics}, also known as the {\em transition model}, are given by 
\begin{equation} \label{eq:transition}
T_a (s, s') = \text{Pr}( s_{k+1} = s'| s_k=s,\ a_k=a)
\end{equation}
which is a probability mass function that leads the agent to succeeding state
$s_{k+1} = s'$ when it executes the action $a_k =a$ from state $s_k=s$. The time step $k$ is also known as the computing epoch, where $1\leq k \leq K$. The fourth element $R_a(s, s')$ is a positive scalar (also called the reward) for performing action $a$ in state $s$ and reaching state $s'$. 
A {\em policy} is a complete mapping $\pi : S \rightarrow A$ so that the agent applies the action $a_k \in A$ in state $s_k \in S$ at step $k$. If the action is independent of $k$, the policy
is called {\em stationary} and $a_k$ is simply denoted by $a$. %, as assumed above. 
}
\end{definition}

\vspace{1pt}
\begin{adjustwidth}{0.3cm}{}
{Note that, unlike conventional formulations, here we use $k$ instead of $t$ to index the algorithmic iterative steps/epochs, for both stationary and non-stationary models. In other words, we regard all $K$ non-stationary algorithm iterations as momentary events, in order to distinguish the temporal process (a.k.a. planning horizon) which will be discussed further in the remainder of this paper.}
\end{adjustwidth}
\vspace{1pt}

\begin{definition}
{\em
The \textit{$\vect{Q}$-value} of a state-action pair $(s, a)$ is defined as the the one-step look-ahead {\em value} of state $s$ if the immediate action $a$ is performed. More formally,
\begin{equation}
Q(s, a) = \sum_{s'\in S}T_a(s, s') \Big( R_a(s, s') + \gamma V (s') \Big),
\end{equation}
where $V(s')$ is the {\em value} (accumulated reward) for state $s'$ and $\gamma\in [0, 1)$ is a discount factor for discounting future rewards at a geometric rate.
}
\end{definition}

The objective is to find an optimal policy $\pi^{*}$ satisfying
\begin{equation}
V_{\pi^{*}}(s) \equiv V^{*}(s) = \max_{a\in A} Q (s, a), \ \forall s\in S.
\end{equation}
When $\gamma < 1$, there exists a stationary policy that is optimal, 
and the optimal policy $\pi^{*}(s)$ is
\begin{equation}\label{eq:policy}
\pi^{*}(s) =\argmax_{a\in A}\!  \sum_{{s'\in S}} T_a(s, s') \Big( R_a(s, s') +\gamma V^{*}(s') \Big)
\end{equation}

Employing Bellman's principle of optimality avoids enumerating solutions naively. 
In particular, the dynamic programming based value iteration (VI) and policy iteration (PI) are the most widely used strategies for solving MDPs~\cite{Puterman94MDP,Bertsekas2000}.

%%%%%%%%%%%%%%%%%%%%%%%%%%%%%%%%%%%%%%%%%%%%%%%%%%%%%%%%%%%%%%%%%%%%%%%%%%%%%%%%%%%%%

\section{Technical Approach}

A conventional MDP has a static transition model $T_a(s, s')$. This means that $T$ has no time-dependence nor does the form of $T$ vary with time$^{\dagger}$.
\footnotetext{$\dagger$ We use {\em time-dependent} and {\em time-varying} to mean different things. Following the convention in operations research, in the MDP context the term {\em time-dependent} is used to describe stochastic travel duration between states, whereas the term {\em time-varying} is used to express the change in the transition model as time elapses.
}
In many practical scenarios,  e.g., marine vehicles in the ocean or aerial vehicles flying in the air, the transition model $T_a(s, s')$ must account for environmental disturbances that vary with time. Such a time variation property requires the control policy to also be a function of time to successfully reject dynamic disturbances and the resultant action uncertainty. 
This motivates us to design a decision-making framework that involves a dynamic and time-varying state transition model.

\subsection{Upgrade from ``Spatial Only" to Both Spatial \& Temporal}

First, the MDP state needs to be upgraded. By ``upgrade" we do not mean to simply append a time variable $t$ to the state variable, e.g., $s:= <s, t>$, so that the classic MDP framework can be directly employed. Such a simple appending operation is problematic because:
\begin{itemize}
    \item Time differs from space in its properties. 
Specifically, space is isotropic and symmetric. Within kinodynamic constraints, we can control an object to freely move in space. In contrast, time is asymmetric and unidirectional.
    \item  The differing properties of space and time imply that,  {\em reachability} in space and time are not the same. We cannot travel back in time.
We have much more restricted reachability (controllability) in time than in space.
\end{itemize}

The above comparison also explains why the standard solution for an MDP cannot be used straightforwardly for an MDP with a time-varying transition model. 
%The standard MDP has no time constraints for state transitions. 
In order to enforce the time constraints, one possible way is to duplicate states into discretized slices of ``time layers",  and then manipulate the transition probabilities so that state hopping backwards is prevented (with a probability of 0).
However, such setting requires a significant increment of state space (sometimes action space also), as well as expensive computational costs for manipulating the transition models.  

In this work, we propose a framework without discretizing time, and thus the state space and action space remain the same as the basic MDP.
We show with a few steps that such spatiotemporal transition model can be addressed using an exact method extended from the MDP, and the state reachability (e.g., states traveling along the time dimension) is explored along algorithmic iterations.

%\vspace{3pt}
\textbf{Correlate Space and Time: }
The difference in reachability in space and time indicates that, 
the state transitions (and value propagation) in space and in time should be treated as two separate processes along a ``spatial channel" and ``temporal channel", respectively. 

The key idea of our work is to {\em evolve} the spatial process of state transition (and value propagation) along the temporal channel, 
where the spatial and temporal processes under two channels are coupled by the underlying real-world physics. For example, in the marine vehicle planning scenario,
the two processes are correlated and coupled by the vehicle's motion dynamics under environmental constraints such as disturbances from ocean currents. 

Formally, to transform an MDP to a TVMDP, the static state transition model $T_a(s, s')$  needs to be a function of time, and we re-write it as $T_a(s, s', t)$. 
Similarly, the reward function becomes $R_a(s, s', t)$. 
The value function is modified accordingly:
\begin{equation} \label{eq:mdp-t} %\vspace*{-3pt}
V^{*}(s, t) = \max_{a\in A} \sum_{{s'\in S}} T_a( s, s', t) \Big( R_a(s, s', t)  + \gamma V^{*}(s', t') \Big),
\end{equation}
where $V(s', t')$ means the value of a future state $s'$ at future time $t'$.
Comparing Eq.~\eqref{eq:mdp-t} with the classic MDP~\eqref{eq:policy}, we see that every term of Eq.~\eqref{eq:mdp-t} is a function of time. 
Thus, the TVMDP can be written as $\mathcal{M}(t) = <S, A, T(t), R(t)>$.
Next, we show how to construct this time-varying decision-making framework in detail.

%\vspace{3pt}
\textbf{Add Real-Valued Transition Time: }
A major difference between the TVMDP $\mathcal{M}(t)$ and the MDP $\mathcal{M}$ lies in that, $\mathcal{M}(t)$ is built on, and thus requires estimation of, the real-valued {\em transition time} for the agent to travel between pairwise states, which is different from the ``state hopping" in an MDP -- an important property inherited from Markov Chains. 

More formally, the {\em transition time} here denotes the observable travel time cost if the agent travels from a state $s$ to another state $s'$. 
It is used to track and map future time moments to their corresponding policy rewards/values at those moments, along the unidirectional temporal channel.

\begin{figure}\vspace{-5pt}
  \centering
  \includegraphics[width=1.8in]{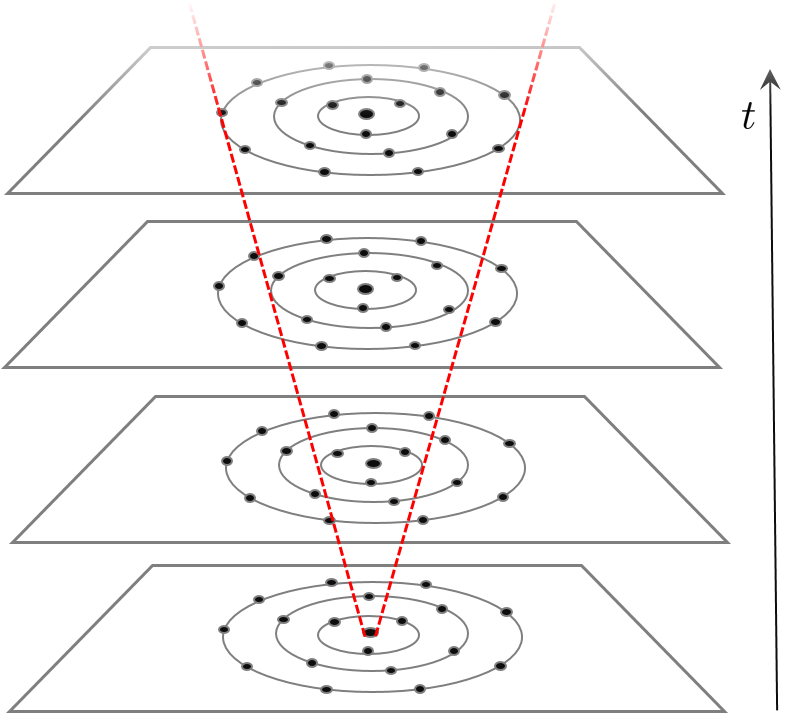} 
 \caption{\small Conceptual illustration of the value iteration propagation along both spatial and temporal dimensions (red dashed lines).
Along the value propagation, the spatial and temporal processes need to be coupled by real world dynamics from both the robot (action/motion) and the environment (disturbance). 
 }\vspace{-10pt}
\label{fig:value-propagation}
\end{figure}

To understand the basic idea, assume that at time $t_0$ the robot is in state $s_0$,  
and let $t(s_0, s)$ be the transition time from $s_0$ to an arbitrary state $s$.
Since the transition model is a function of time and the time elapses during the robot's motion,  
thus when the robot eventually arrives at state $s$, the transition model that impacts the robot is given by $T_a(s, s', \tau)$ 
instead of $T_a(s_0, s)$ (or equivalently $T_a(s_0, s, t_0)$),  where $\tau=t_0 + t(s_0, s)$ is a later time. 
%(Here $s'$ is the previously defined succeeding state that $s$ can directly transit to.)  
Therefore, value evaluation/iteration for state $s$ at the starting moment $t_0$ needs to look ahead and utilize the transition model $T_a(s, s', \tau)$ captured at the later time $\tau$.
Fig.~\ref{fig:value-propagation} depicts such an idea.
It is worth mentioning that, the discrete time layers in Fig.~\ref{fig:value-propagation} are used only for the purpose of demonstrating the idea of an additional time dimension. We are showing that the proposed method does not need to discretize time.

Also note that, here we use a perfect prediction model that assumes the estimated times $t(s_0, s)$ and $\tau$ are accurate. This however, is unrealistic due to the robot's stochastic behavior. This problem is discussed and addressed in Sect.~\ref{sect:trans-time}.

%%%%%%%%%%%%%%%%%%%%%%%%%%%%%%%%%%%%%%%%%%%%%%%%%%%%%%%%%%%%%%%%%%%%%%%%%%%%%%%%%%%%%

%\vspace{5pt}
\textbf{Construct Time-Varying Transition Models: }
\label{sec:tran-model}
The mapping of future time moments to future policy values essentially relies on employing the correct transition models at those future moments.
This is because policy values are propagated on the stochastic network that is expressed by those transition models.
Hence, {\em the key component of a TVMDP is the time-varying transition model}.
By a time-varying transition model, we mean that the state transition distribution is time-varying (e.g., it can be caused by the dynamic disturbances of the environment).
This differentiates our proposed framework from the time-dependent formulations~\cite{Boyan00exactsolutions}, in which the transition model is used to represent the stochastic travel duration between pairwise states.   

Time-varying transition model can be predicted by utilizing the forecast environmental dynamics. E.g., the forecast data of ocean currents in next few days can define a predictable planning horizon, within which the extracted disturbance dynamics can be used to predict the robot's future state transition properties.

Specifically, since both the robot's action $a$ and the external disturbance $d$ are usually two independent forms of sources that cause the robot's (uncertain) motion, similar to the transition distribution $T_a(s, s')$ caused by the action $a$, we use $T_d(s, s')$ to represent the transition distribution caused by the pure disturbance $d$.
The synergistic transition model $T_{ad}(s, s')$ caused by both the action $a$ and the disturbance $d$ can be obtained by unifying their two independent transition distributions,  and all those distributions can be associated with time $t$ if the action/disturbance is a function of time.  Formally,
\begin{equation}\label{eq:synergistic}
T_{ad}(s, s', t) = T_a(s, s', t) \oplus T_d(s, s', t)
\end{equation}
where the operator $\oplus$ ``combines" two distributions captured (or predicted) at moment $t$. The operations of combining probability distributions will depend on specific distribution types, e.g., see~\cite{ranjan2010combining,winkler1981combining}. 
In our robotics motion planning scenario, we assume both $T_a(s, s', t)$ and $T_d(s, s', t)$ follow independent Gaussian distributions. 

\subsection{Estimation of Transition Time}
\label{sect:trans-time}

Before estimating the time-varying transition models corresponding to the future moments, the real-valued transition times traveling from current state to all other states need to be estimated first.
Intuitively, the transition time estimation is a process of ``look ahead into future".
%based on which the transition dynamics can then be predicted. 
One might also imagine this as a {\em forward} process, as time evolves in a forward direction.

However, such a forward estimation cannot be done in a straightforward way. 
This is because the forward process/propagation is usually used on a deterministic acyclic graph, but in the MDP state transition topology, there are both cycles and stochasticity.
The estimation from one state to another actually relies on each other, resulting in a chicken and egg problem. 
In addition, we need to estimate not only the time of one-step look-ahead, but also those times many steps ahead corresponding to all other distant states within the planning horizon. 

Our approach consists of two steps. The first step is to estimate ``local" transition time from each state $s\in S$ to its {\em one-hop} succeeding states $\mathcal{N}(s)\subseteq S$ that $s$ can directly transit to. Building on this, the second step is to estimate the ``global" transition time from the robot's current state $s_0$ to all other {\em multi-hop} states $S\setminus \mathcal{N}(s_0)$. Details are as follows.

\vspace{3pt}
\textbf{Local One-Hop Transition Time Estimate:}
%\vspace{1pt}
The concept of one-hop can also be interpreted as one-step look ahead, since it measures the agent's travel time from current state to the one-hop adjacent neighboring states.
Note, the one-hop time estimate is {\em domain-specific} and may be done in multiple ways. 
For example, with a known transition model that essentially describes the state hopping probability distribution from the current state to neighboring states, the one-hop time can be tested and estimated offline by Monte Carlo trials.
There are also other closed-form solutions by taking advantage of the known state transition probability distributions, %however, due the limitation of space,  the details will not be discussed in this paper. 
and an application example in the marine robotic planning scenario is provided in the Appendix.

\vspace{3pt}
\textbf{Global Multi-Hop Transition Time Estimate: }
%\vspace{1pt}
Using the local one-hop transition time, 
we can then estimate the global multi-hop transition time $t(s, s_e)$ from an arbitrary state $s$ to an end state $s_e\in S$ which does not immediately succeed $s$.
Multi-hop transition time estimation is {\em domain-independent}.

Estimating the global multi-hop transition time is challenging because
the time estimate from an arbitrary state $s$ to another multi-hop state $s_e$ needs to take into account many combinations of possible hopping scenarios due to the underlying cyclic nature. This causes interrelated dependence among states: estimation of arrival time at $s_e$ travelled from $s$ relies on a starting time at $s$, which essentially relies on the time estimates of all other states that directly or indirectly connect to $s$, including the state $s_e$.  

One possible solution is to compute the {\em first passage/hitting} time at state $s_e$. 
However, the classic {\em Markov Chain first-passage times} method~\cite{siegert1951first} does not apply here because it only considers and analyzes, the number of hops, instead of estimating the real-valued travel time.

Our solution is to formulate the problem using Kolmogorov equations~\cite{kolmogorov31}. For example, from $s_1$, the transition time $t(s_1, s_e)$ can be represented by an expectation $\vect{E}\left(t(s_1, s_j^{(1)}) + t(s_j^{(1)}, s_e) \right)$, where $s_j^{(1)}\in \mathcal{N}(s_1)$ is $s_1$'s one-hop succeeding states, and $t(s_j^{(1)}, s_e)$ is again a multi-hop transition time from $s_j^{(1)}$ to ending state $s_e$. 
Similarly, we can formulate an expression for the transition time $t(s_i, s_e)$ for all other $s_i \in S$ to $s_e$:
\begin{equation*}\label{eq:kolmogorov}
\left\lbrace %
\begin{split}
    t(s_1, s_e) =\mkern-18mu \sum_{{s_j^{(1)} \in \mathcal{N}(s_1)} } & \mkern-18mu T_a(s_1, s_j^{(1)}, t)\Big( t(s_1, s_j^{(1)}) + t(s_j^{(1)}, s_e) \Big)\\
            &\setbox0\hbox{=}\mathrel{\makebox[\wd0]{\vdots}} \\
 t(s_n, s_e) = \mkern-18mu \sum_{{s_j^{(n)} \in \mathcal{N}(s_n)} } &\mkern-18mu T_a(s_n, s_j^{(n)}, t) \Big(t(s_n, s_j^{(n)}) + t(s_j^{(n)}, s_e) \Big)
\end{split} %
\right.
\end{equation*}
where $t(s_e, s_e) = 0$ and each $t(s_i, s_j^{(i)})$ denotes the previously obtained one-hop transition time.

We have a total of $|S|$ variables and $|S|$ equations, which form a linear system. 
From the current state $s_0$, there are  $|S\setminus \mathcal{N}(s_0)|$ multi-hop non-succeeding states. Therefore, we need to solve 
$|S\setminus \mathcal{N}(s_0)|$ linear systems each of which specifies a different end state $s_e \in S\setminus \mathcal{N}(s_0)$.
%In many decision-making scenarios, the number of one-hop succeeding states are  limited, thus many coefficients in the linear system are zeros, which can be solved by sparse linear system solvers (with complexity O(|S|^{2.3})).
Solving (sparse) linear systems is the most time expensive part during each value iteration, and it requires a complexity of $O(|S|^{2.3})$~\cite{Golub1996}. 
Thus the overall time complexity to compute all $|S|$ estimates is $O(|S|*|S|^{2.3})$ for each iteration.

\subsection{Synergy of Spatial and Temporal Processes}

The final step to solve the proposed TVMDP is to propagate policy values in order to maximize optimal actions. 

After having obtained the transition time  $t$ from the robot's current state $s_0$ to an arbitrary state $s$,
the transition model at state $s$ and time $t$ can be constructed:% by combining the environment disturbance information~\eqref{eq:transition-t}, i.e., 
\begin{equation}
T_{ad}(s, s', t)  = T_{ad}(s, s', t(s_0, s)).
\end{equation}
where $s'\in S$ are the possible successor states from $s$.

In contrast to the forward time estimation process, the value propagation is more like a {\em backward} process:
the propagation process employs a Bellman backup mechanism and propagates the rewards from distant states in the future back to the current state at the current moment. 
To improve estimation accuracy, the forward time estimate process and backward value propagation process need to alternate iteratively until a certain (preset) convergence threshold is met.

After each iteration of Bellman backup, the action policy is updated. 
Based on the updated policy, the transition time estimates from the current state to all other states $t(s_0, s)$ $\forall{s\in S}$ are calculated, which are then used to update the time-varying transition model $T_{ad}(s, s', t(s_0, s))$ at $s$. The updated transition model is in turn utilized for next value iteration/propagation and time estimate. 
The corresponding pseudo-code is described in Alg.~\ref{algo:VI}.
In essence, the underlying computing mechanism can be imagined as value iteration that combines both spatial ``expansion" and temporal ``evolution", which occur simultaneously in two separate  channels. This is a solution mechanism that fundamentally differs from that of the classic MDP. In fact, the MDP's Bellman backup procedure can be regarded as the {\em spatial expansion} process, and thus the MDP is a special case of TVMDP. 

The proposed time estimation formulations produce unique solutions (solving them does not involve iterative optimization processes). Thus after we integrated the time estimations with the MDP value iterations, the convergence behavior mainly depends on the standard value iterations.
The algorithm will converge because along the algorithmic stages, both the time estimates and state values (updated by value iterations) are constantly improved and become more and more accurate. 
Since  value iteration is known to be near-optimal depending on the convergence stopping tolerance, the solution to TVMDP is thus expected to be near-optimal depending on the accuracy of time estimation. Formally, the solution can become ``suboptimal" only when the time estimates are inaccurate compared to ground truth (e.g., local time estimate model is imperfect, forecast data and dynamics are inaccurate, vehicle state and control are noisy, see Appendix).

\begin{algorithm}
\caption{Value Iteration (with agent's state $s_0$)}
\label{algo:VI}
Initialize: $k\leftarrow 1$, \\
\ForEach{$s \in S$}{
Initialize $V_0(s) = 0$, $t(s_0, s)=0$
}
\vspace{3pt}
\ \ /* {\em value propagation in spatial channel} */\\
\ForEach{$s\in S$}{ 
$\pi_k^{*}(s) = \texttt{argmax}_{a\in A} \sum_{s'\in \mathcal{N}(s)} T_a(s, s', t(s_0, s))\cdot$ \\
$\quad \quad \big( R_a(s, s', t(s_0, s)) + \gamma V_{k-1} (s', t(s_0, s')) \big)$ \\
update optimal action $a^{*}(s) = \pi_k^{*}(s)$
}
\vspace{3pt}
\ \ /* {\em transition time estimates in temporal channel} */\\
\ForEach{$s\in S$}{ 
\ForEach{$s_j \in \mathcal{N}(s)$}{
Estimate one-hop transition time $t(s, s_j)$
}
}\ForEach{$s\in S\setminus \mathcal{N}(s_0)$}{ 
Estimate multi-hop transition time $t(s_0, s)$
}
$k\leftarrow k+1$, \textbf{goto} Step 5\\
Terminate algorithm if given tolerance is reached\\
%\line(1,0){100} \\
%{\scriptsize {\bf Note:} } }
\end{algorithm} 

%%%%%%%%%%%%%%%%%%%%%%%%%%%%%%%%%%%%%%%%%%%%%%%%%%%%%%%%%%%%%%%%%%%%%%%%%%%%%%%%%%%%%

%\setlength{\belowcaptionskip}{-10pt}

%\vspace{-10pt}
\section{Experiments}
\label{sec:expt}

\begin{figure} [t]  \vspace{-5pt}
  \centering
  \subfigure[]
        {\label{fig:policy-surface}\includegraphics[height=1.7in]{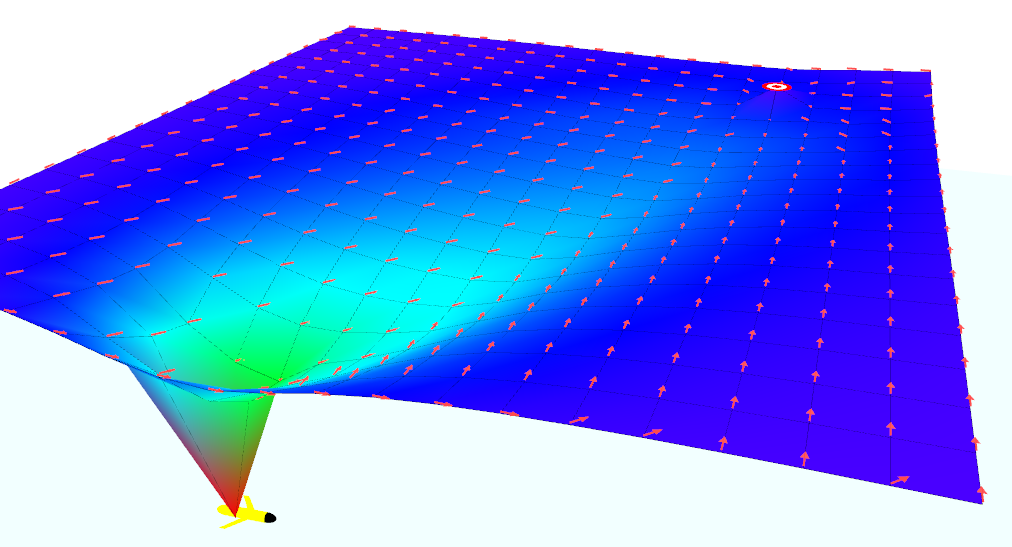}}   
  \subfigure[]
        {\label{fig:region-sigma1}\includegraphics[height=1.2in]{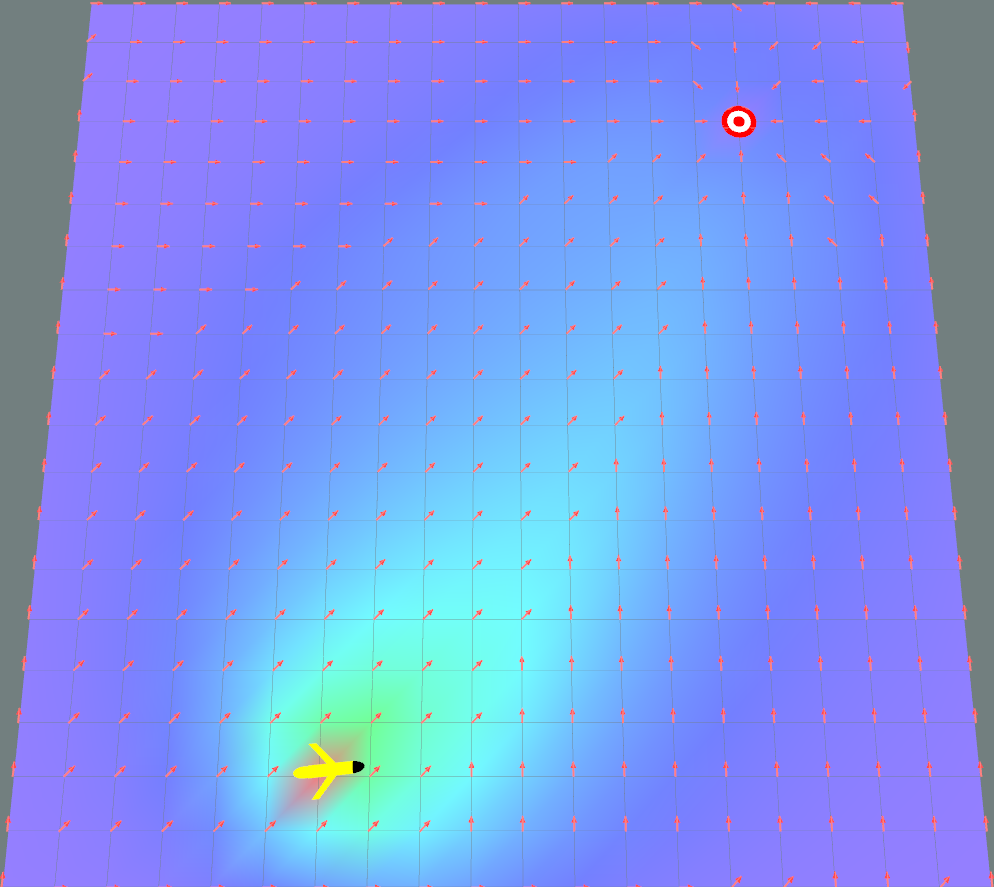}}
        \quad
  \subfigure[]
        {\label{fig:region-sigma2}\includegraphics[height=1.2in]{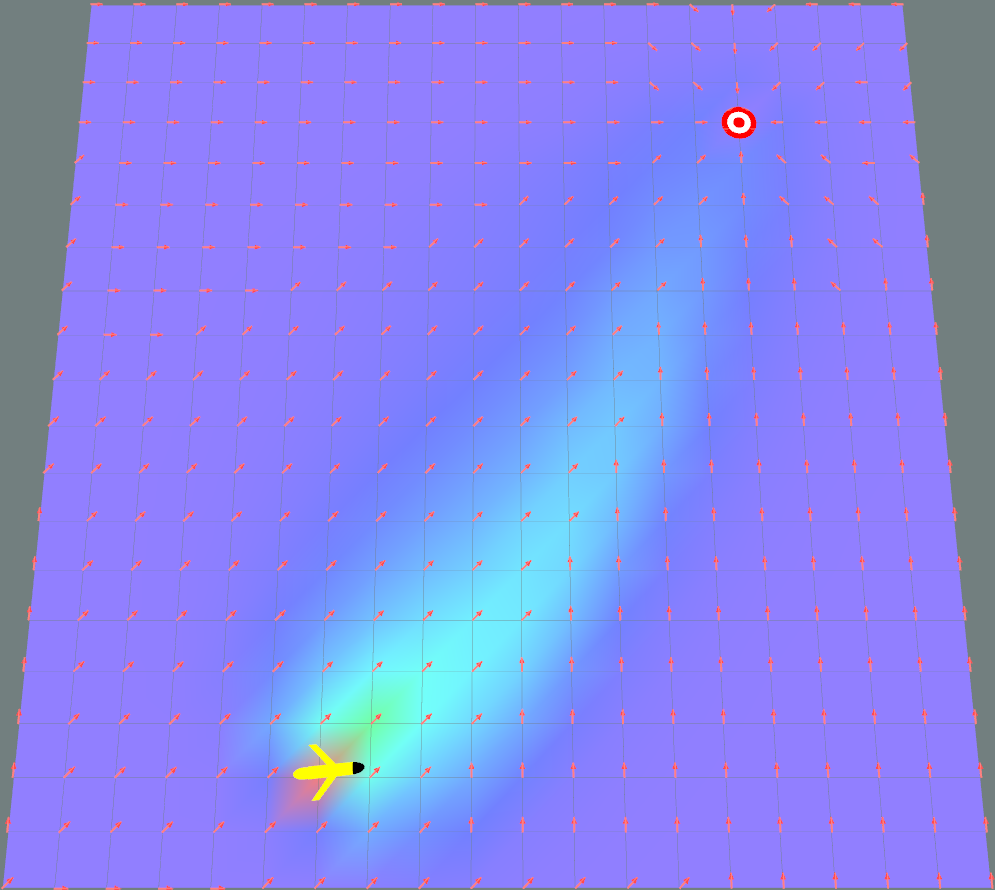}}
  \caption{\small (a) TVMDP action policy (red arrows)  on a funnel-like surface where the height of funnel represents estimated transition time from the robot's state (funnel bottom); (b)(c) Policy maps projected onto 2D spatial dimensions. The colormap implies reachability in space. Different magnitudes of action uncertainty result in differing areas of reachable regions. The red target symbol is the goal state to reach. }
\label{fig:fig:action-policy}   \vspace{-10pt}
\end{figure}

We validated our method in an ocean monitoring scenario, where the ocean currents vary both spatially and temporally.
An underwater glider simulator written in C++ was built in order to test the proposed decision-making framework.
%The robot used in simulation is a underwater glider with a simplified kinematic model. 
The simulation environment was constructed as a two dimensional ocean surface, and we tessellated the environment into a grid map. 
We represent the center of each cell/grid as a state, where each non-boundary state can transit in eight directions (N, NE, E, SE, S, SW, W, NW) and a ninth idle action (returning to itself).
Time-varying ocean currents are external disturbances for the robot and are represented as a vector field. 

We first investigated the policy patterns generated from the proposed framework with time-varying transition models. 
%While considering only robot action and its uncertainty (no external disturbance),
Fig.~\ref{fig:policy-surface} shows a policy map (red arrows) on a  funnel-like surface where the height of the funnel is a measure of estimated transition time  from the robot state (bottom of the funnel). 
Fig.~\ref{fig:region-sigma1} and~\ref{fig:region-sigma2} are projected policy maps onto a 2D plane. A brighter region implies a larger chance of being visited by the robot, and the difference of brighter regions in two figures reveals differing ``magnitudes" of action uncertainty (Fig.~\ref{fig:region-sigma1} has larger action uncertainty than Fig.~\ref{fig:region-sigma2}). 
   
\begin{figure*}  %\vspace{-10pt}
  \centering
   \subfigure[MDP]
        {\label{fig:mdp_}\includegraphics[height=1.in]{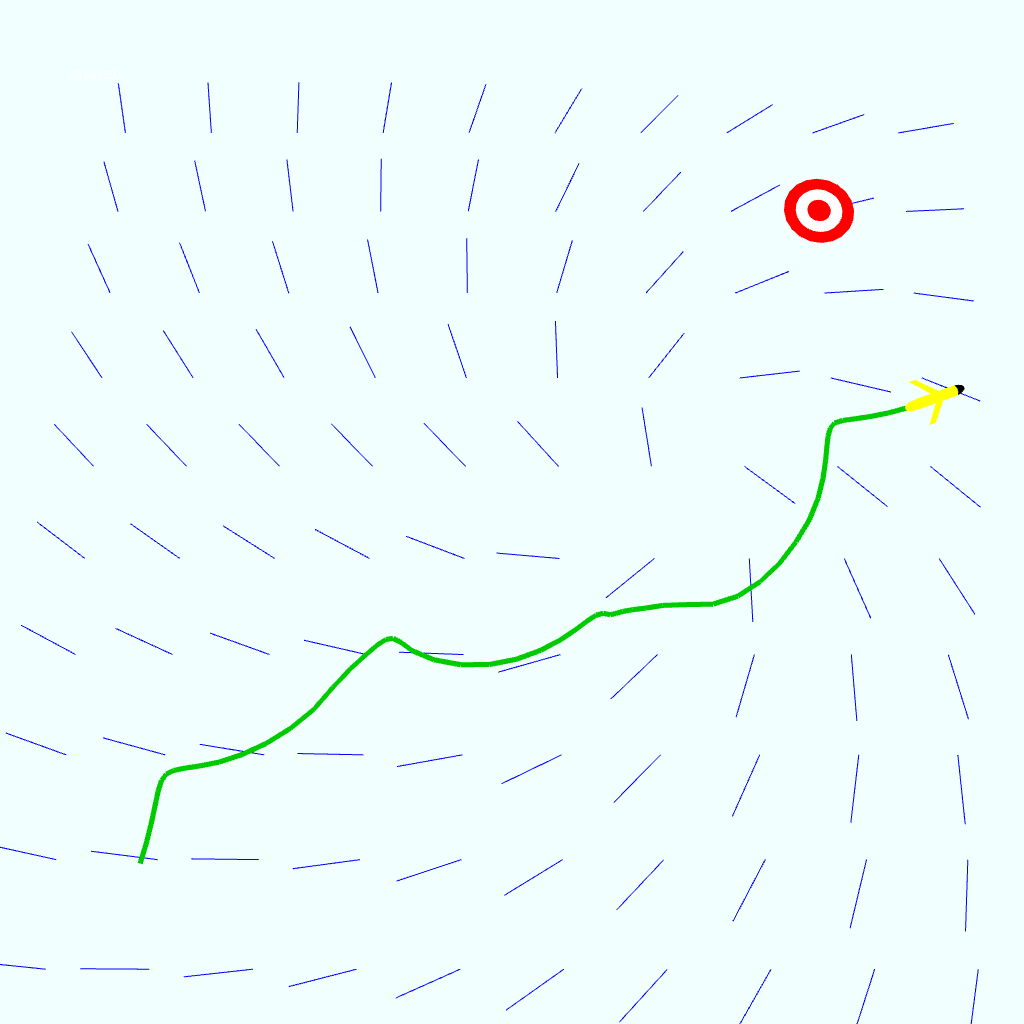}}
        \quad 
  \subfigure[DTMDP low]
        {\label{fig:dtmdp1}\includegraphics[height=1.in]{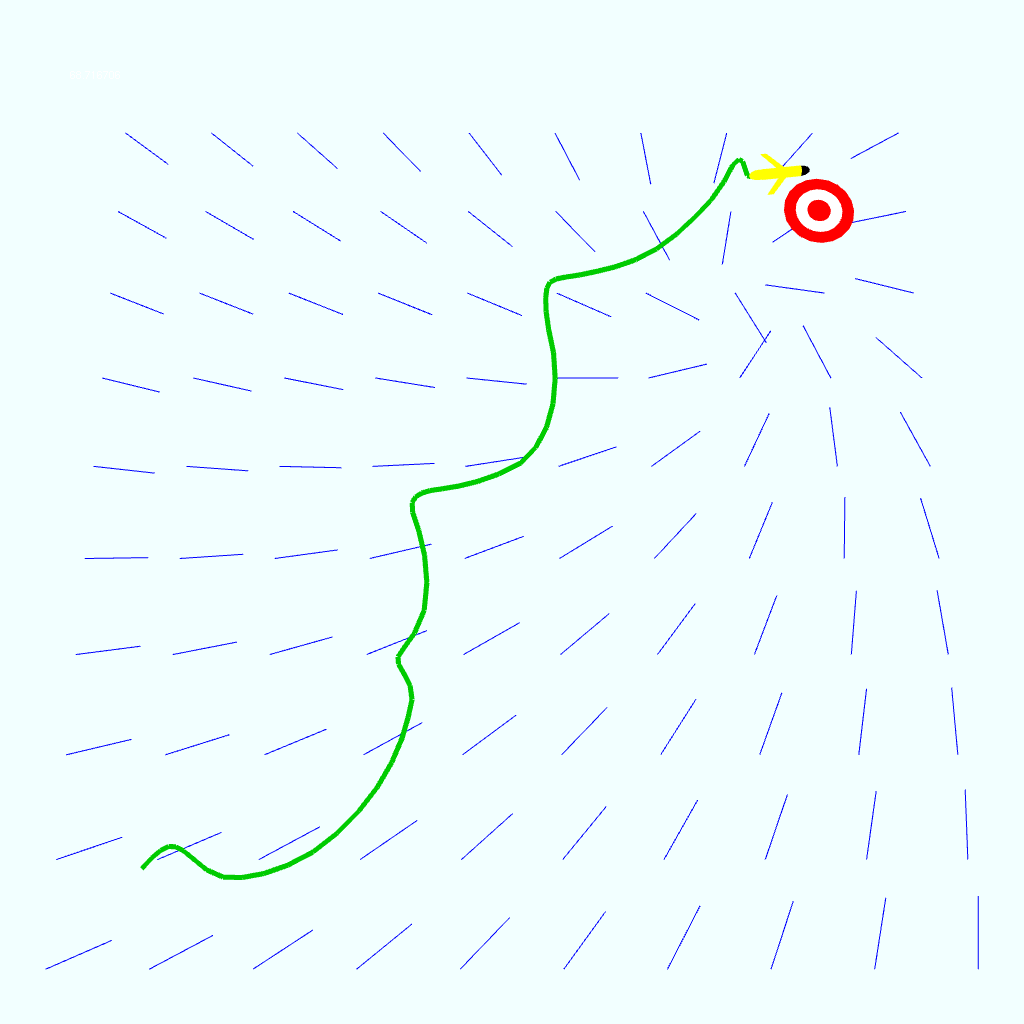}}
        \quad 
  \subfigure[DTMDP high]
        {\label{fig:dtmdp2}\includegraphics[height=1.in]{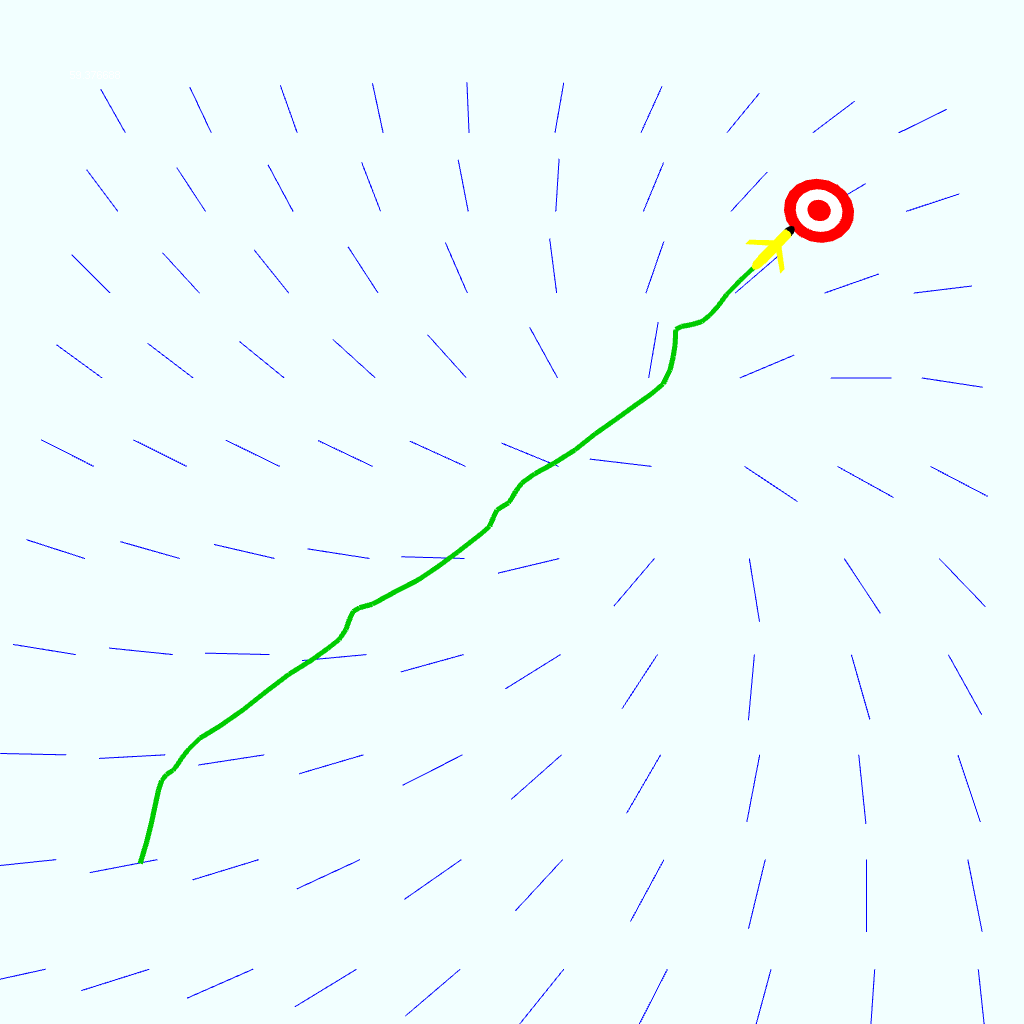}}
        \quad
  \subfigure[TVMDP]
        {\label{fig:tvmdp_}\includegraphics[height=1.in]{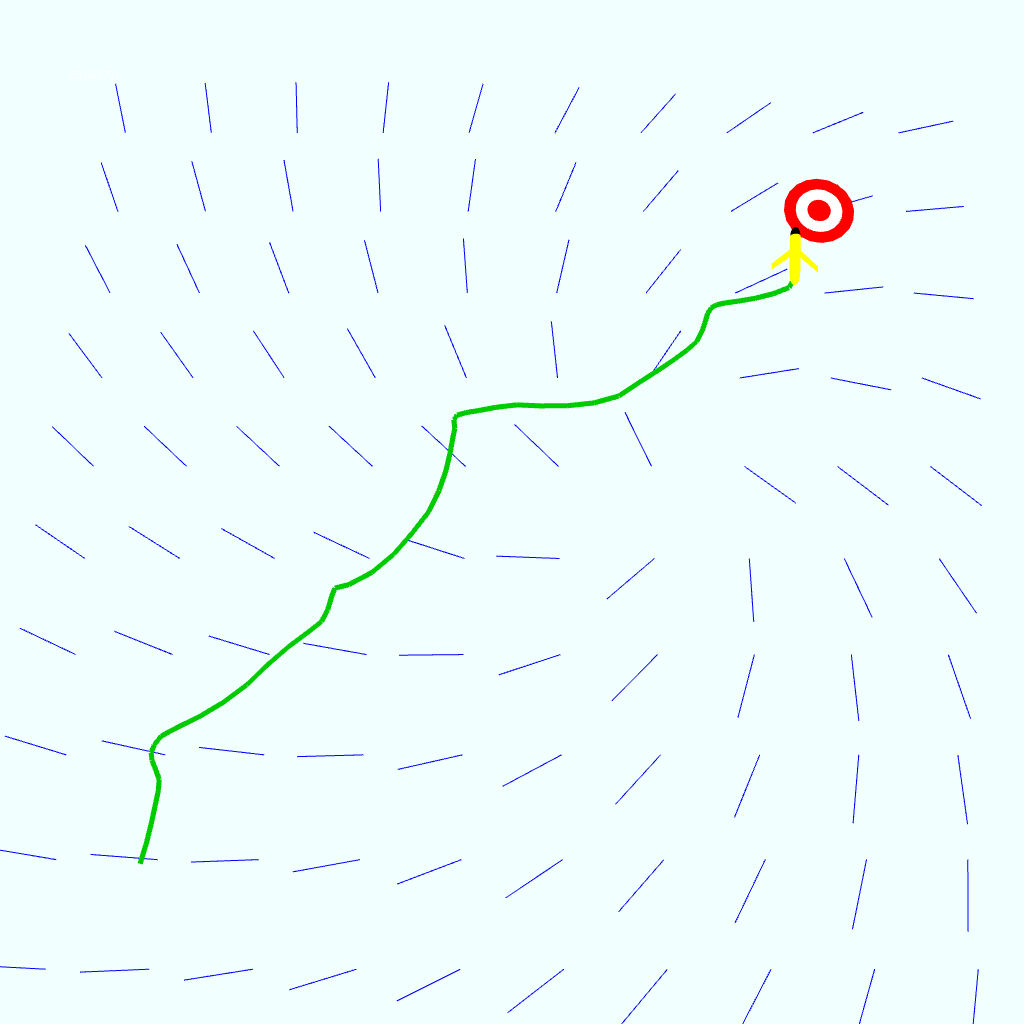}} \vspace{-5pt}\\
  \subfigure[ATMDP]
        {\label{fig:tdmdp-spin}\includegraphics[height=1.1in]{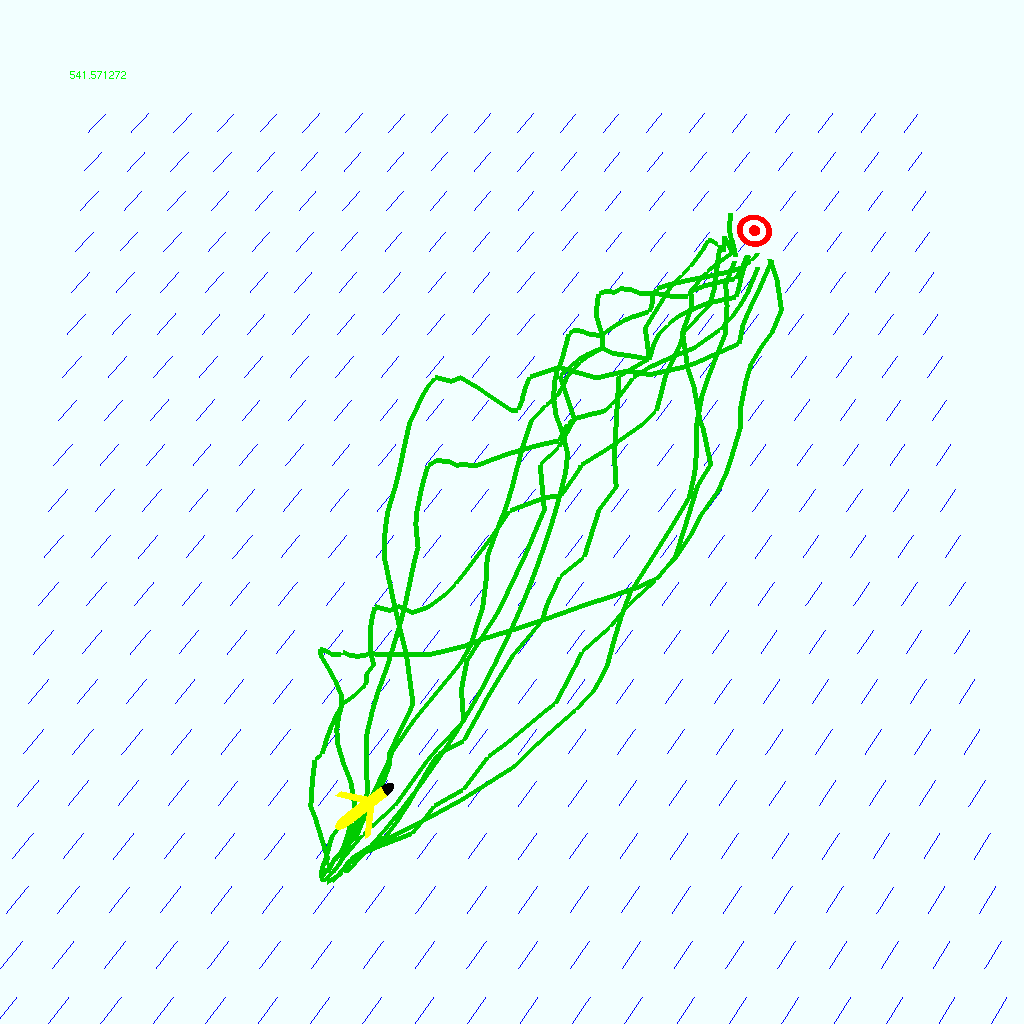}}
        \quad 
  \subfigure[TVMDP]
        {\label{fig:tvmdp-spin}\includegraphics[height=1.1in]{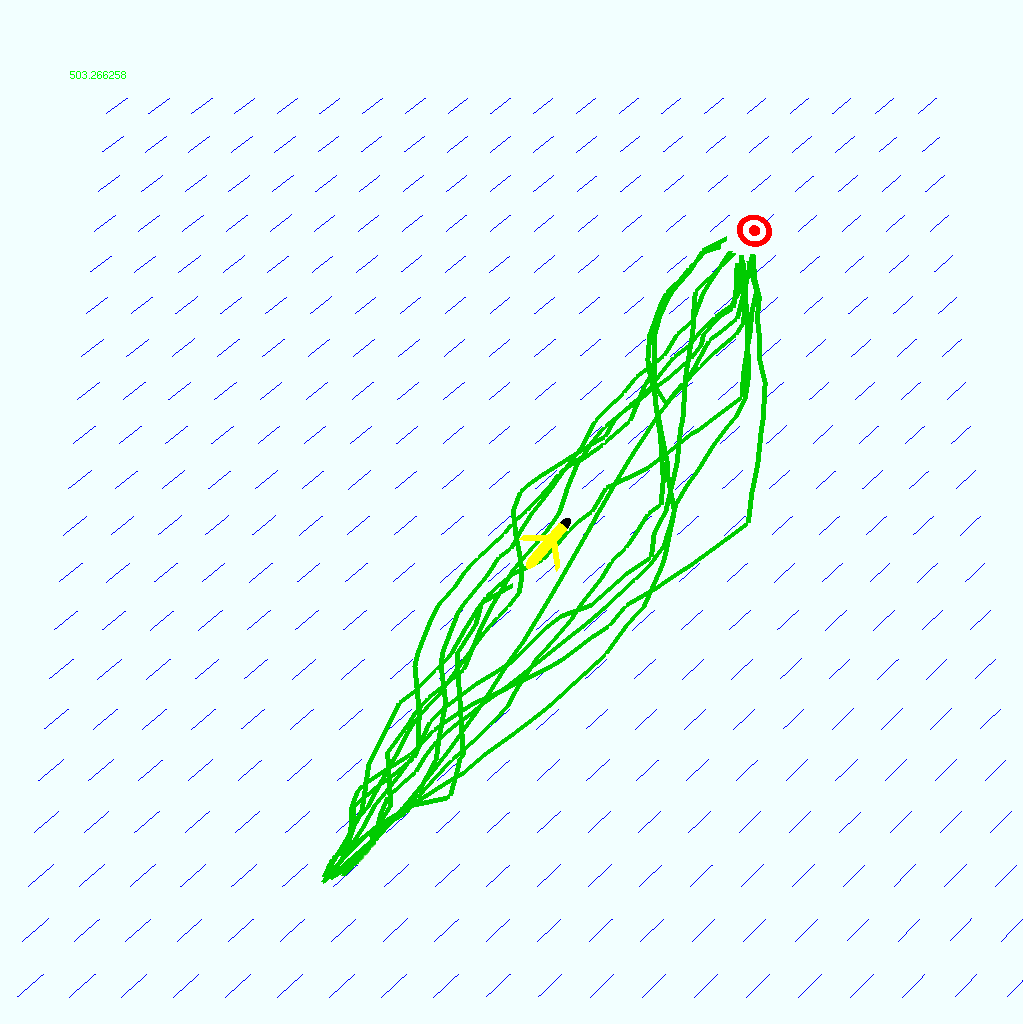}}
        \quad 
  \subfigure[ATMDP]
        {\label{fig:tdmdp-vortex}\includegraphics[height=1.1in]{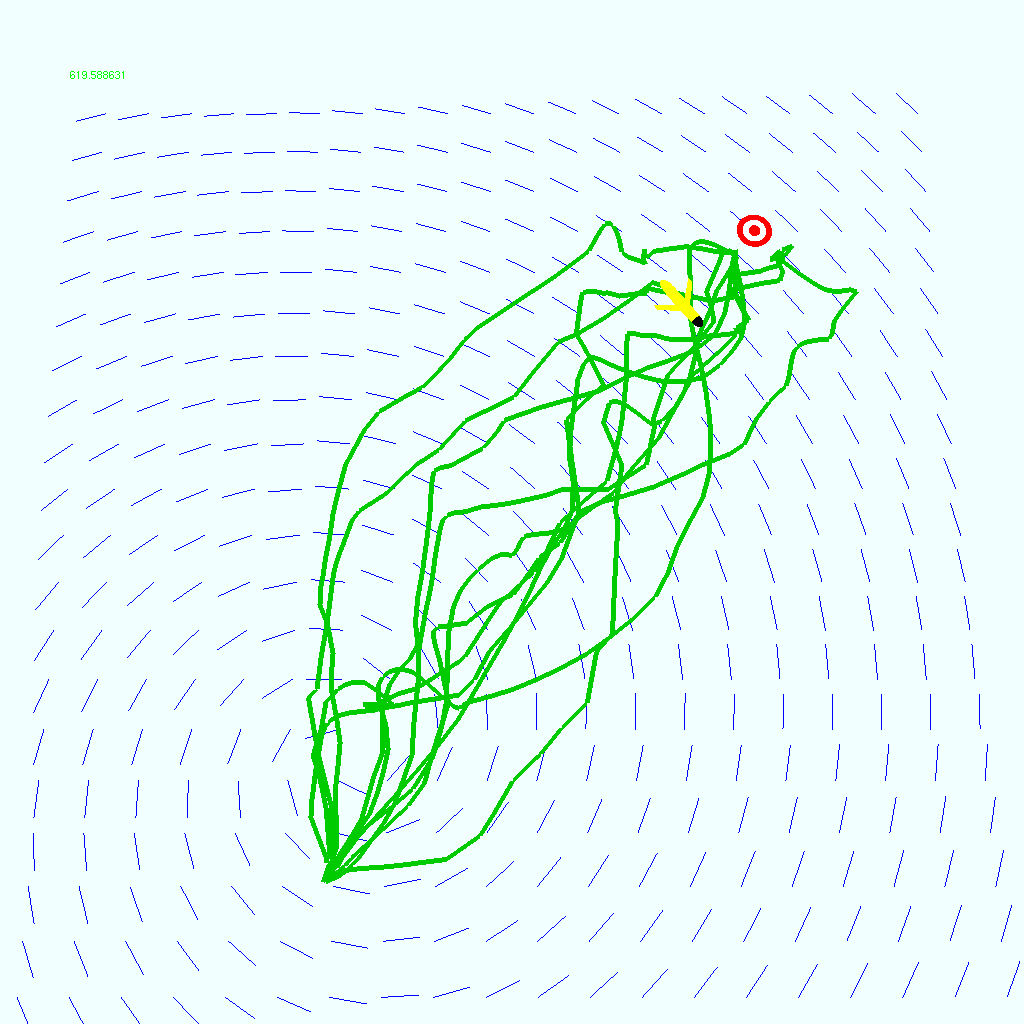}}
        \quad
  \subfigure[TVMDP]
        {\label{fig:tvmdp-vortex}\includegraphics[height=1.1in]{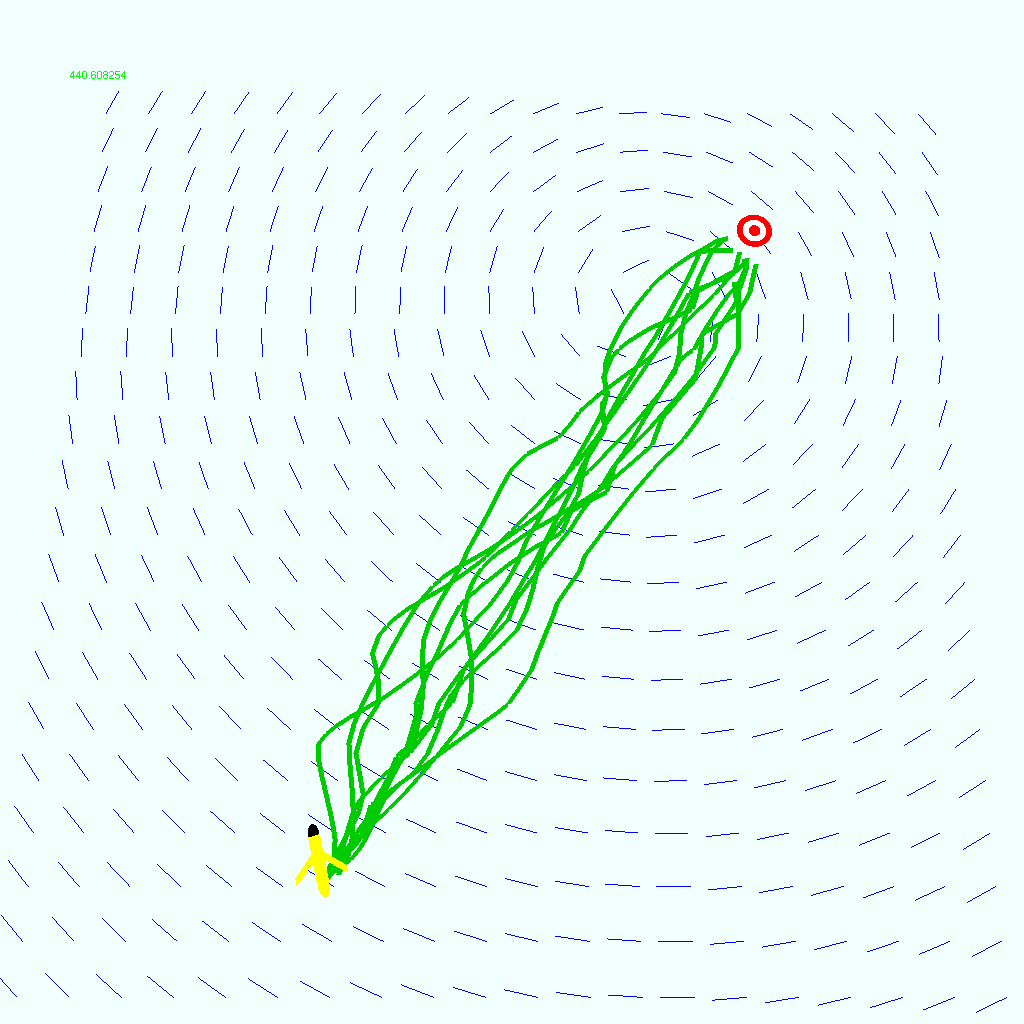}}
 \caption{\small The top four pictures demonstrate trajectories from (a) MDP; (b) DTMDP with low time-discretization resolution; (c) DTMDP with high time-discretization resolution; (d) TVMDP.  The disturbance is a dynamic vortex-like vector field.  
The bottom four pictures show accumulated trajectories of many trials resulted from ATMDP and TVMDP, in spinning and vortex vector fileds, respectively.
 %(a)(b)(e)(f) Robot trajectories under a spinning disturbance vector field $\vect{d}([x, y]_t) = R(\omega t)[d, d]^T$,  where $R(\omega t)$ is the rotation matrix with rotating rate $\omega$ and $d$ is a constant value.;\ \ (c)(d)(g)(h) Trajectories under a vortex-like vector field $\vect{d}([x, y]_t) = [-\big(x-x_c(t)\big) + \big(y-y_c(t)\big), -\big(x-x_c(t)\big)-\big(y-y_c(t)\big)]$ where $[x_c(t), y_c(t)])$ is a dynamic vortex center.
  }
  \label{fig:traj-compare}   %\vspace{-10pt}
\end{figure*}

To evaluate the proposed method, we compare it with other relevant approaches including the standard MDP, the discrete-time MDP (DTMDP), as well as a modified version of time-dependent MDP (TDMDP)~\cite{Boyan00exactsolutions} since it shares certain similarity with our framework.

\begin{figure}
  \centering 
  \subfigure[]
        {\label{fig:cost-dt}\includegraphics[height=1.35in]{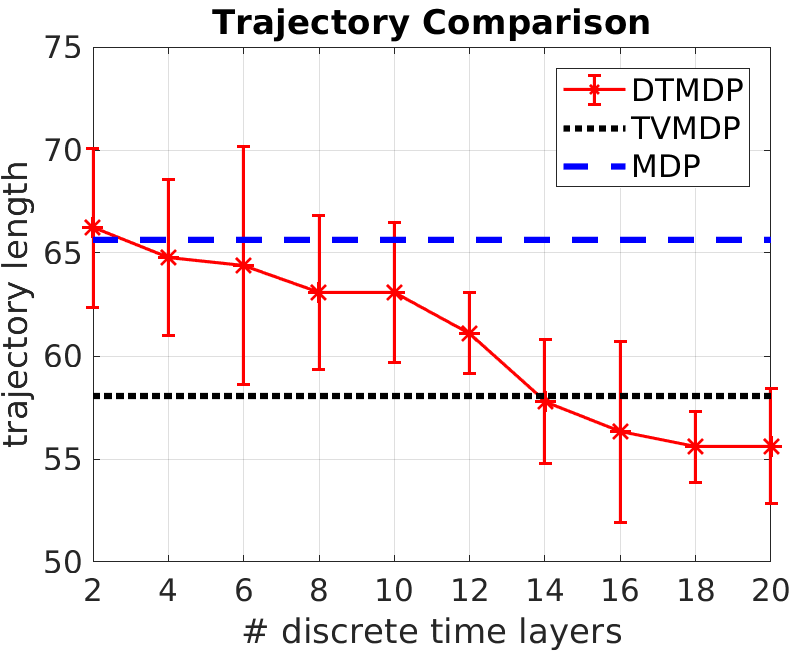}}
        \
  \subfigure[]
        {\label{fig:time-dt}\includegraphics[height=1.35in]{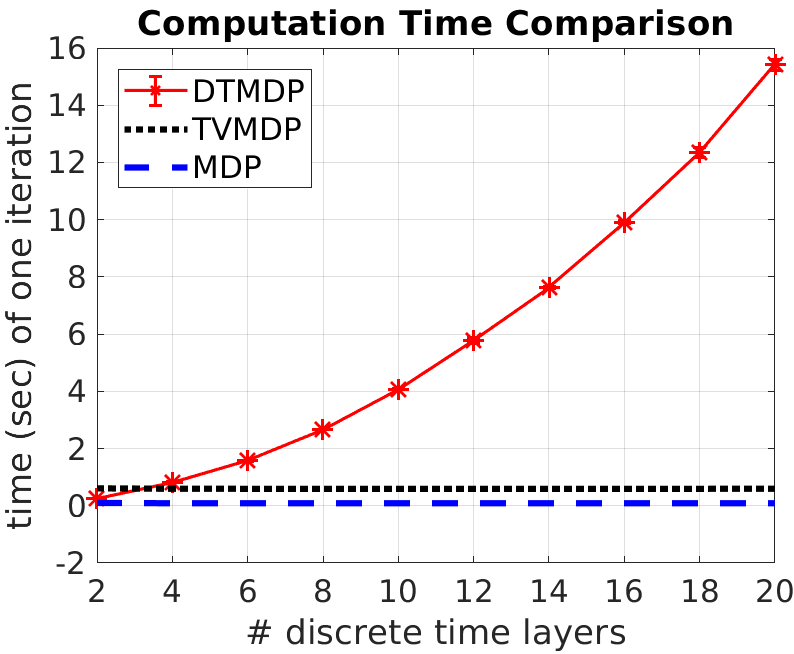}}\vspace{-10pt}
  \caption{\small Performance comparisons between TVMDP, MDP, and DTMDP. The two dashed lines are averaged values from MDP and TVMDP, respectively. The grid map has a dimension of $10\times10$, thus the number of states is 100 for MDP and TVMDP, and is $100d_t$ for DTMDP where $dt$ is the number of time layers.
  }
\label{fig:statistics-cost-time}  \vspace{-10pt}
\end{figure}

We start by comparing TVMDP with MDP and DTMDP. 
The DTMDP extends from MDP by adding a time dimension, i.e.,  duplicating MDP states into a series of discretized time layers. This method relies on a pre-defined time horizon as well as discretization resolution. 
Fig.~\ref{fig:mdp_}- \ref{fig:tvmdp_} demonstrate trajectories from these methods under a dynamic vortex-like disturbance field. We can observe that, DTMDP is very sensitive to the resolution of time discretization. Specifically, the method generates a good result when the resolution is high enough, but produces a bad performance when the resolution is low.
Statistics in Fig.~\ref{fig:statistics-cost-time} reveal more details. 
We can see from Fig.~\ref{fig:cost-dt} that, the trajectory length/cost of DTMDP has an obvious decreasing trend as the number of time layers  (time-discretization resolution) grows, and can actually result in a better performance than TVMDP if the time resolution is high enough.
The reason for the sub-optimality of TVMDP is due to imperfect time estimates (see approximation method in Appendix).
%and the estimation error grows over time.
Note, however, an important drawback of DTMDP lies in its prohibitive computational cost. Fig.~\ref{fig:time-dt} shows computational costs for a $10\times10$ grid map and only one value iteration. This implies that the prohibitive cost of DTMDP makes it less useful for most of application scenarios. 

We also compared TVMDP with a modified version of time-dependent MDP (TDMDP) and we call it {\em approximate time-varying MDP (ATMDP)}.
(The TDMDP is not straightforwardly applicable in our scenario due to the reasons we analyzed earlier. We modified it by employing an approximation method of {\em prioritized sweeping}~\cite{RachelsonFG09} which essentially propagates the largest value function changes in priority through the state space, such that the dilemmas caused by the stochastic and cyclic topology are mitigated and a look-ahead time estimate solution is approximated.)
Fig.~\ref{fig:tdmdp-spin}--\ref{fig:tvmdp-vortex} show trajectories produced from the ATMDP and TVMDP. We can see that, the trajectories of our method are much smoother and shorter.

\begin{figure}  [t]  
  \centering
  \subfigure[]
        {\label{fig:path}\includegraphics[height=1.3in]{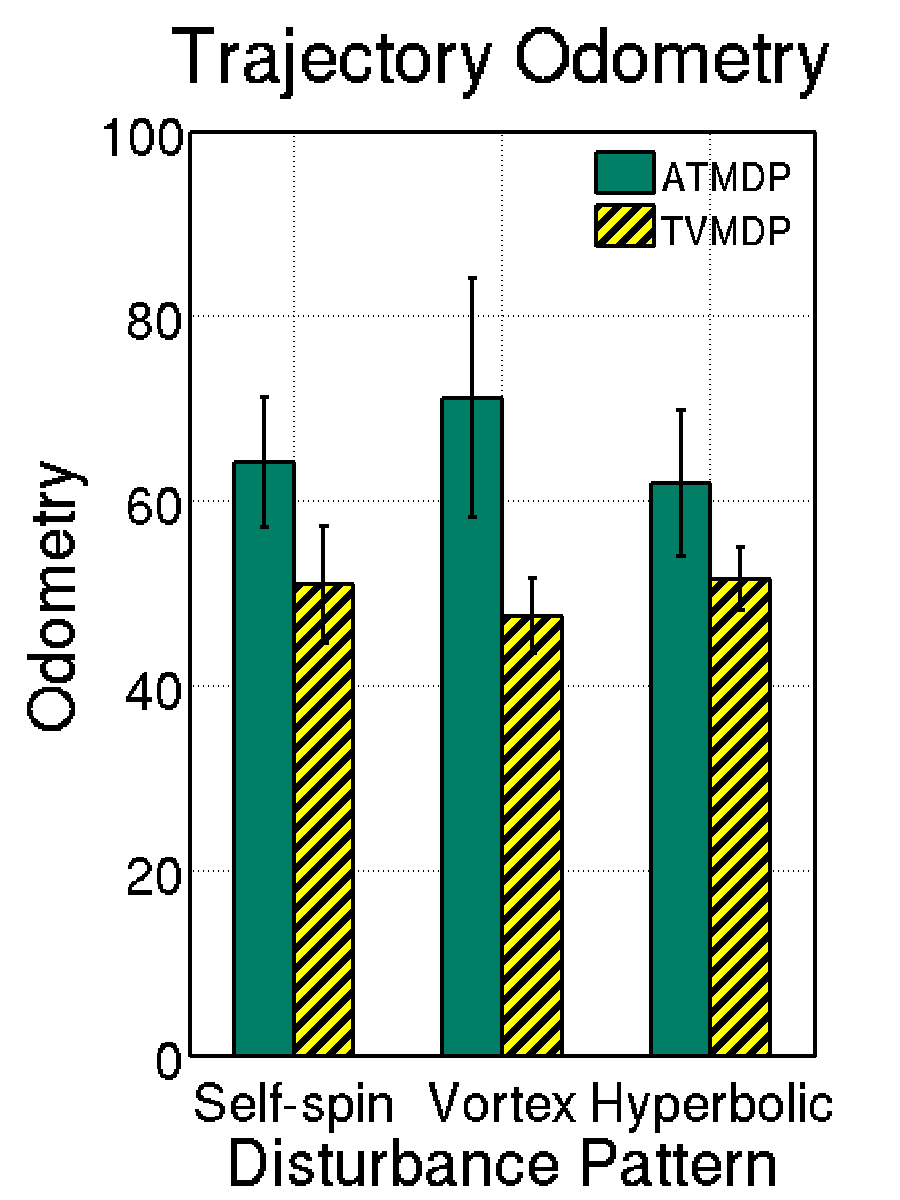}}
  \subfigure[]
        {\label{fig:time}\includegraphics[height=1.3in]{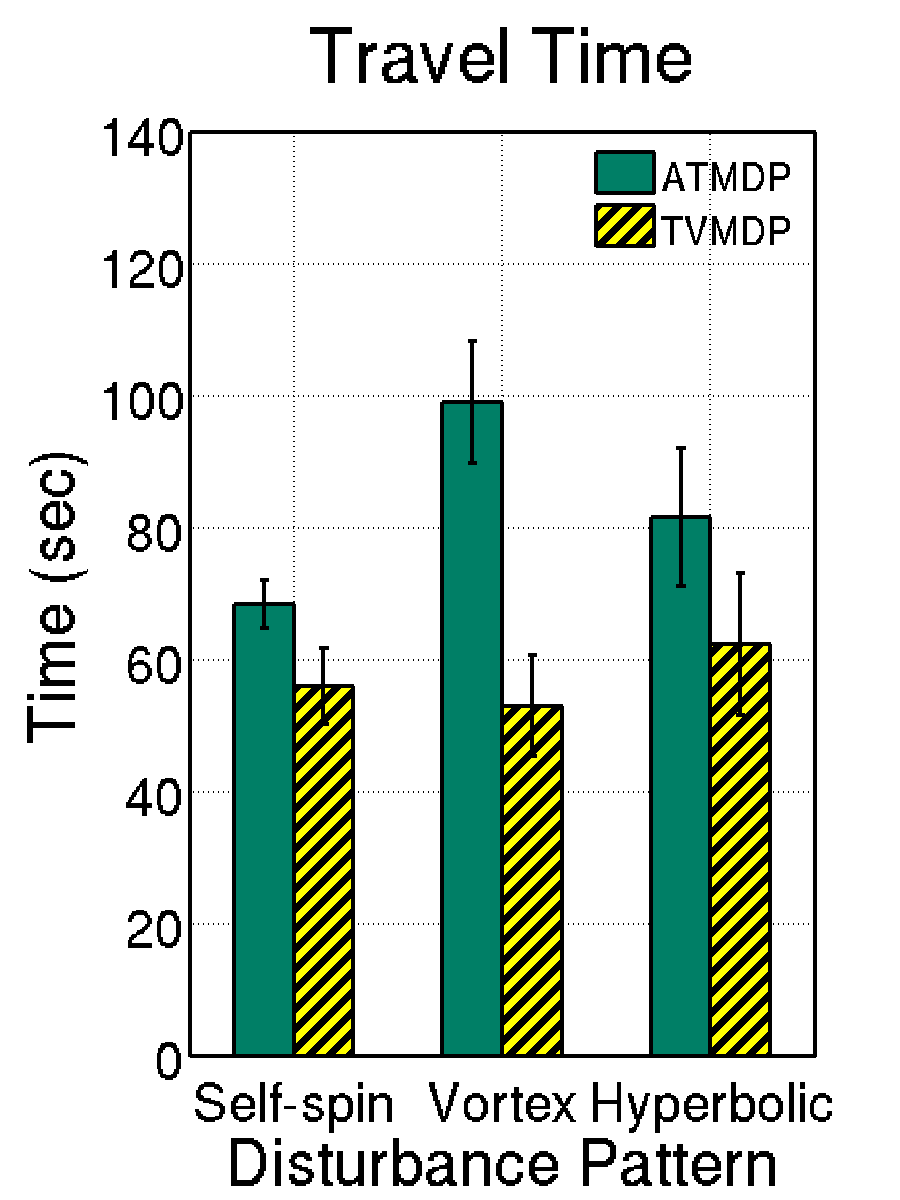}}
  \subfigure[]
        {\label{fig:compute}\includegraphics[height=1.3in]{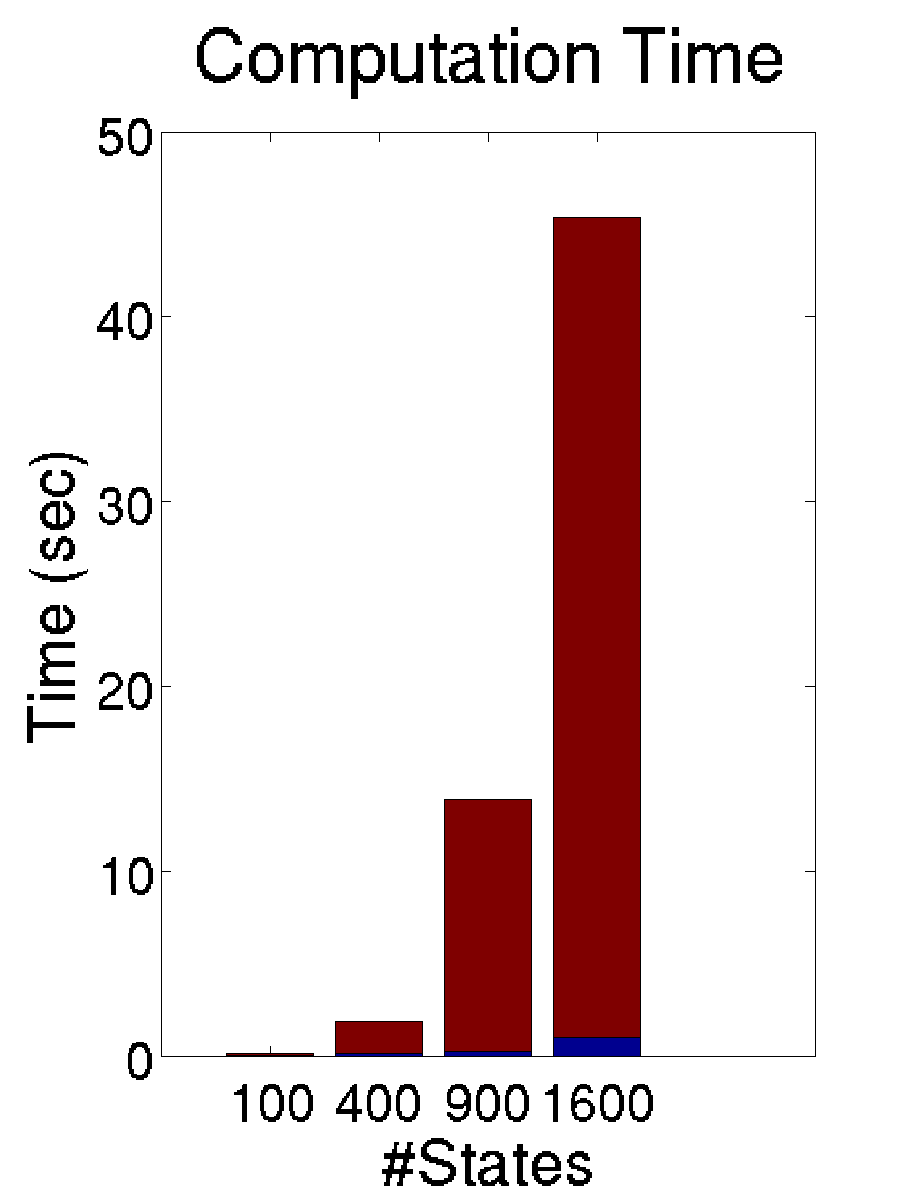}}
  \caption{\small Performance comparisons between ATMDP and the TVMDP, under differing disturbance patterns. (a) Trajectory odometry (lengths);  (b) Overall travel time; (c) Computational time required by TVMDP to generate solutions.
%The blue stack at the bottom of each bar denotes the time used for value iterations of standard MDP. 
The red parts are the time used for solving linear systems. 
}
\label{fig:statistics}  \vspace{-10pt}
\end{figure}

\begin{figure}
  \centering 
  \subfigure[ATMDP]
        {\label{fig:atmdp-ocean}\includegraphics[height=1.55in]{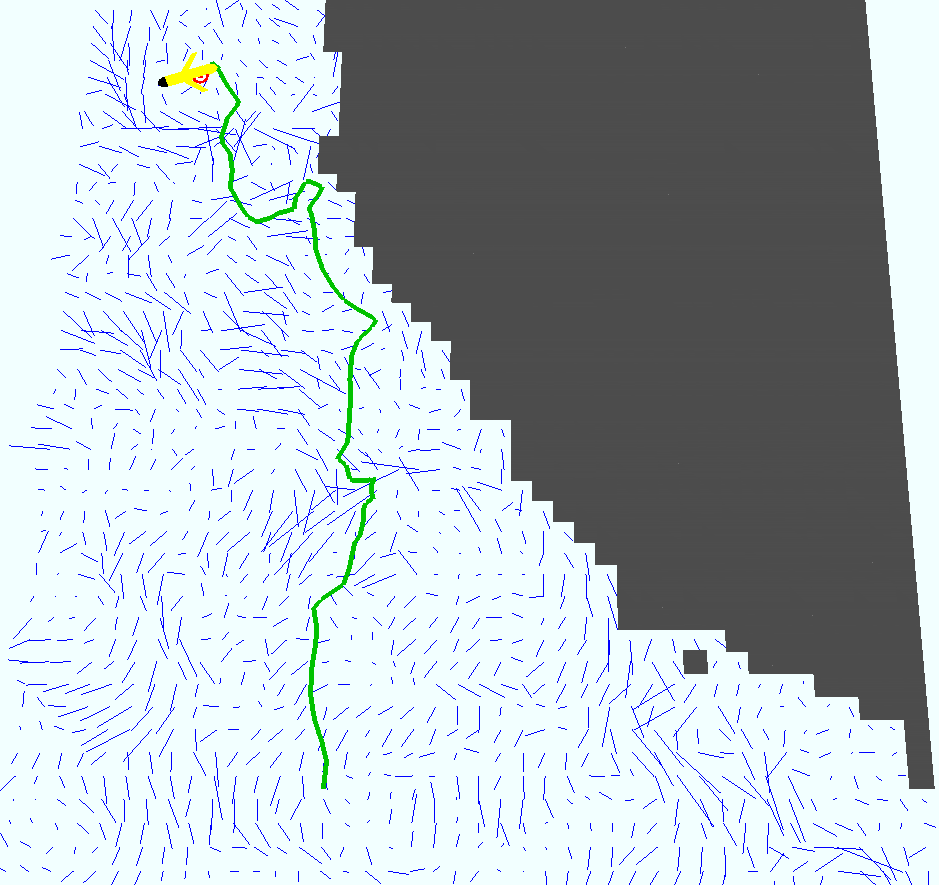}}
  \subfigure[TVMDP]
        {\label{fig:tvmdp-ocean}\includegraphics[height=1.55in]{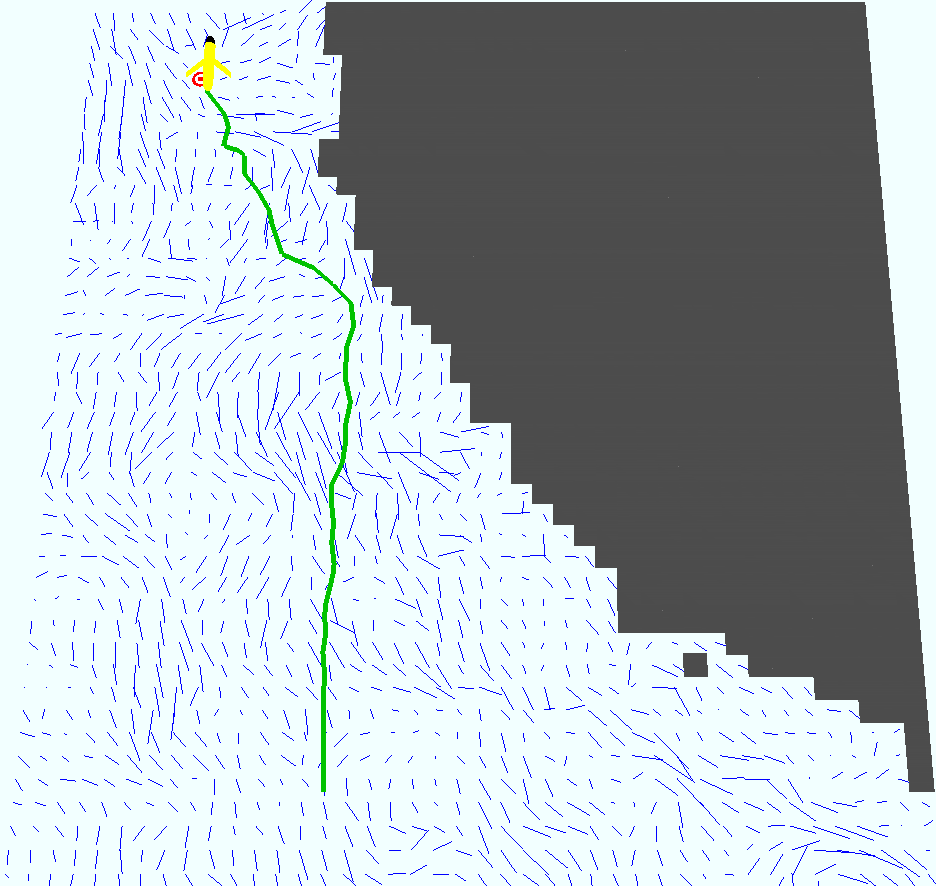}}
  \caption{\small Trajectory results from running on ROMS data.  (a) Trajectory from ATMDP; (b) Trajectory from TVMDP.
  %With our method, the robot needs shorter time and travel distance to reach a specified goal location. 
  }
\label{fig:roms}  \vspace{-10pt}
\end{figure}

Statistics with regard to trajectory lengths and time costs for TVMDP and ATMDP are provided in Fig.~\ref{fig:path} and~\ref{fig:time}. These results indicate that the TVMDP method leads to smaller travel distances and shorter travel times. We use the \texttt{Eigen} iterative sparse matrix solver to compute linear systems for estimating multi-hop transition time. Fig.~\ref{fig:compute} shows that our method requires $\sim$15 seconds for $\sim$1000 states and $\sim$50 seconds for $\sim$ 2000 states (on a desktop with 3.30GHz Intel $i7$ CPU).   

Finally, we also tested the algorithms using real ocean current data.
The simulator is able to read and process data from the 
Regional Ocean Model System (ROMS)~\cite{shchepetkin_regional_2005} which predicts/forecasts ocean currents up to 72 hours in advance. 
This allows us to utilize these ocean predictions to model the temporal dimensional transition dynamics.
%The ocean currents are labelled with latitude $lat$, longitude
%$lon$, the current easting component $u$ and northing component $v$, as well the time $t$. 
However, ROMS provides eight datasets for one day (every three hours). This means the data is not time-continuous. To address this, we use Gaussian Processes Regression (GPR) to interpolate and extrapolate the missing parts. 
Fig.~\ref{fig:roms} shows results from running the ATMDP and the TVMDP. 
%Since the task spans multiple days, we set the prediction horizon of TVMDP as three hours, i.e., whenever new prediction data is available. 
We can observe that, comparing with TVMDP, only the beginning part of ATMDP trajectory is good. This is because ATMDP uses an approximation method that can only provide very rough time estimation results, and the approximation errors are propagated as the horizon grows, which eventually leads to poor policies for the later part of the trajectory.

%%%%%%%%%%%%%%%%%%%%%%%%%%%%%%%%%%%%%%%%%%%%%%%%%%%%%%%%%%%%%%%%%%%%%%%%%%%%%%%%%%%%%

%\vspace{-10pt}
\section{Conclusion}
%\vspace{-5pt}

We presented a time-varying MDP framework called TVMDP which is able to handle transition models that vary both spatially and temporally.
We show that in order to solve the TVMDP, the basic Bellman backup mechanism used for computing the MDP is not enough. Instead, 
we developed mechanisms to estimate the temporal parameter as well as integrate the time-varying stochasticity. Our proposed mechanism consists of two iterative value propagation processes that proceed in both spatial and temporal dimensions, where the spatial process involves Bellman backup but the temporal dimension is expanded using Kolmogorov equations.
Finally we validated our method with various dynamic disturbances including those from real ocean data. 
The results show that comparing to the conventional methods, our approach produces high quality solution with a low computational cost.

\vspace{5pt}
\appendix\label{appendix}
\noindent\textbf{A Marine Robotic Motion Planning Example}

The purpose of this appendix is to demonstrate a possible way of estimating the real-valued one-hop transition time described in Section~\ref{sect:trans-time}. 

Since the estimation of any real-valued transition time is based on continuous state space,  
we first need to convert the conventional discrete MDP state space to a continuous one. 
There are many methods to do so (e.g., \cite{toussaint2006probabilistic,sanner2012symbolic}).
To facilitate the description of the main algorithm, we adopt a simple strategy that maps from/to the other within certain resolution. 
The continuous form of action and external disturbance are also defined accordingly:

\begin{itemize}
\item State $\vect{x}$ is the counterpart of MDP state $s$ but in continuous state space. When $\vect{x}$ coincides exactly at $s$, we denote it as $\vect{x}(s)$ in continuous space. We can also map $\vect{x}$ back to discrete space: $\vect{x} \mapsto s$ if $||\vect{x} - \vect{x}(s)||$ is less than discrete space partition resolution;
\item Local control/action reference $\vect{a} (s)$ at $s$ is a directional vector, where the vector direction is defined by the discrete action $a \in A$, and the vector magnitude is defined by the robot's actual control inputs.
\item Vector $\vect{d} (s)$ expresses external disturbance at $s$.
\end{itemize}

Consequently,  if the transition from state $s$ to a succeeding state $s'$ is deterministic, the transition time can be approximated by $|| \vect{x}(s) - \vect{x}(s') || / v$, or simply $|| \vect{x} - \vect{x}' || / v$, assuming the robot's speed $v$ is a constant in a static environment.

We assume the outcomes of $\vect{a}(s)$ and $\vect{d}(s)$ are stochastic and specifically they follow independent Gaussian distributions.  
After a travel time $\mathcal{T}$, the robot's movement translation $\vect{x}$ (denoting arriving state) after applying $\vect{a}(s)$ and being disturbed by $\vect{d}(s)$ also follows a  Gaussian distribution:
\begin{equation}\label{eq:normal}
f_a(\vect{x}) = \frac{1}{\sqrt{(2\pi)^d|\Sigma|} } \text{exp} \left(-\frac{1}{2} (\vect{x}-\vect{\mu})^T \Sigma^{-1}(\vect{x}-\vect{\mu}) \right)
\end{equation}
where $\mu$ and $\Sigma$ are the mean and covariance of $\vect{x}$, respectively.
It is worth mentioning that the MDP may produce multiple optimal actions (with equal optimal value) at some state. In such a case, a {\em mixture distribution} can be used,
\begin{equation}
f_{\{a_1,\cdots, a_k\} }(\vect{x}) = \sum_{i=1}^{k} w_i f_{a_i}(\vect{x}) = \frac{1}{k}\sum_{i=1}^{k} f_{a_i}(\vect{x}),
\end{equation}
where the weighting parameter $w_i$ for component PDFs are identical as actions have the same optimal value.

Let $\{\vect{a}^{*}(s)\}$ be the set of optimal actions at state $s$ and at time $t$, i.e., $s_t$ for short, the time-varying transition model thus can be expressed as
\begin{equation} \label{eq:transition-t}
T_{ad} (s, s', t) =   
%\text{Pr}\big( s_t' \mid s_t, \vect{r}_t(s)\big) =
\text{Pr}\big( s' \mid s_t, \{\vect{a}^{*}(s_t)\}, \vect{d}(s_t) \big).
\end{equation}
In practice, such discrete probability mass function is approximated by integrating Eq.~\eqref{eq:normal} over discretized volumes.

%\textbf{Local One-Hop Transition Time Estimate}

With the above model, we can proceed to compute the local one-hop transition time estimate. 

Due to the stochastic nature of the transition model in Eq.~\eqref{eq:transition-t}, the robot in state $s$ may eventually arrive at any one-hop succeeding state $s'\in \mathcal{N}(s)$. 
However, the estimation of transition time $t(s, s')$ is based on the assumption that the robot will reach a designated next state $s'$ with probability 1. To satisfy this  assumption, we need to choose an action $\tilde{\vect{a}}(s)$ with which the robot motion is exactly toward $s'$. Let $\tilde{\vect{r}}(s) = \tilde{\vect{a}}(s)+\vect{d}(s)$ denote the resultant of such selected action and environmental disturbance at $s$. 
%and $\vect{x}'$ be the resultant continuous state due to $\vect{r}'(s)$. 

To simplify the calculation, one way is to transform the coordinate system such that the robot's motion direction $\frac{\tilde{\vect{r}}(s)}{||\tilde{\vect{r}}(s)||}$ is exactly on an arbitrary coordinate basis, built on which the multivariate PDF (Eq.~\eqref{eq:transition-t}) can be approximated by a univariate PDF by marginalization.

\begin{figure} \vspace{-5pt}
  \centering
  \includegraphics[height=1.6in]{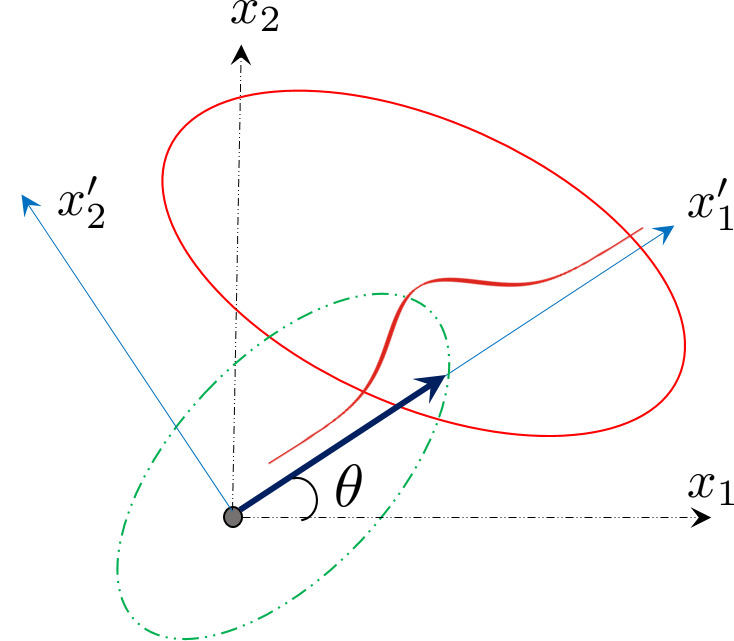} 
 \caption{\small Transition time estimation in 2D. The black dot at the origin represents the robot's current state $s$; the red solid ellipse denotes the state continuous distribution $f(\vect{x})$ under $\tilde{\vect{r}}(s) = \tilde{\vect{a}}(s) + \vect{d}(s)$ (the bold black arrow); the green dashed ellipse represents a contour of $\vect{r}(s)=\vect{a}(s) + \vect{d}(s)$ for all allowable $\vect{a}(s)$ given that $\vect{d}(s)$ is fixed.
 }
\label{fig:local-time} \vspace{-10pt}
\end{figure}

Fig.~\ref{fig:local-time} shows an illustration with robot state $s$ located at the coordinate origin. 
%The red solid ellipse denotes the state continuous distribution $f(\vect{x})$ under $\tilde{\vect{r}}(s) = \tilde{\vect{a}}(s) + \vect{d}(s)$ (the bold black arrow); and the green dashed ellipse represents a contour of $\vect{r}(s)=\vect{a}(s) + \vect{d}(s)$ for all allowable $\vect{a}(s)$ given that $\vect{d}(s)$ is fixed.
$\theta$ is the angle between $\tilde{\vect{r}}(s)$ and a basis $x_1$. To transform the coordinate system such that $\tilde{\vect{r}}(s)$ is on $x_1$, the coordinate needs to rotate with corresponding rotation matrix $R(\theta)$. More generally, with a rotation matrix $R$ and replacing $\vect{x} = R \tilde{\vect{x}}$ in Eq.~\eqref{eq:normal}, we obtain a distribution in the transformed coordinate system: 
\begin{equation*}
\begin{split}
f(\tilde{\vect{x}}) 
&\propto \text{exp} \left(-\frac{1}{2} (\tilde{\vect{x}}-R^{-1}\vect{\mu})^T R^T \Sigma^{-1}  R(\tilde{\vect{x}} - R^{-1}\vect{\mu})\right)\\
%&\propto \text{exp} \left(-\frac{1}{2} (\tilde{\vect{x}}-R^{-1}\vect{\mu})^T (R^T \Sigma R)^{-1} (\tilde{\vect{x}}-R^{-1}\vect{\mu})\right)\\
&\propto\text{exp} \left(-\frac{1}{2} (\tilde{\vect{x}}-R^T\vect{\mu})^T (\tilde{\Sigma})^{-1} (\tilde{\vect{x}}-R^T\vect{\mu})\right)
\end{split}
\end{equation*}
where $\tilde{\mu} = R^T\vect{\mu}$ and $\tilde{\Sigma} = R^T \Sigma R$ after transformation.
 
Next we can calculate the {\em expectation} of resultant states in the robot's motion direction. Let $x_i$ denote a selected $i$-th basis (variable) in the direction of motion in the {\em new coordinate system}, and $\vect{x}_{-i} = (x_1, \cdots, x_{i-1}, x_{i+1}, \cdots, x_d)$ represent all other variables.
Then the conditional expectation of $x_i$ can be calculated by 
\begin{equation*}
\begin{split}
\vect{E}(x_i | \vect{x}_{-i} &= \vect{0}) = \int x_i f(x_i | \vect{x}_{-i}) I_{\{\vect{x}_{-i} = \vect{0}\}} (\vect{x}_{-i} ) \text{d} \vect{x}\\
& = \int x_i \frac{f(x_i, \vect{x}_{-i})}{f(\vect{x}_{-i})} I_{\{\vect{x}_{-i} = \vect{0}\}} (\vect{x}_{-i} ) \text{d} \vect{x}\\
& = \int_0^{\tau} \int x_i \frac{f(x_i, \vect{x}_{-i})}{f(\vect{x}_{-i})} I_{\{\vect{x}_{-i} = \vect{0}\}} (\vect{x}_{-i} ) \text{d} \vect{x}_{-i}\text{d} x_i
\end{split}
\end{equation*}
where indicator variable $I_{\{\vect{x}_{-i} =\vect{0}\}} (\vect{x}_{-i}) = 1$ if $\vect{x}_{-i} = \vect{0}$ and $0$ otherwise. 
%This reduces the multivariate distribution to a one dimensional distribution on axis of $x_i$. 
$\tau$ can be either $+\infty$ (motion in the direction of $x_i$) or $-\infty$ (motion against the direction of $x_i$).
%The denominator $f(\vect{x}_{-k})$ can be calculated by first marginalizing $x_i$.

Since states $\tilde{\vect{x}}$ are obtained by applying $\tilde{\vect{r}}(s)$ for a fixed time $\mathcal{T}$, the local one-hop transition time $t(s, s')$ is approximated by 
\begin{equation}
t(s, s') = \frac{|| \vect{x} - \vect{x}'||}{\left(\vect{E}(x_i | \vect{x}_{-i} = \vect{0})/\mathcal{T}\right)}
\end{equation}
where $|| \vect{x} - \vect{x}' ||$ represents the translation defined as before  
%(Euclidean distance) between the two continuous states 
and the denominator is the estimated velocity given that the destination state is $s'$.

%\end{comment}

%\vspace*{-2pt}
{\small
\bibliographystyle{abbrv}
\bibliography{reference}
}

\end{document}